\documentclass[10pt,twocolumn,letterpaper]{article}

\usepackage{cvpr}              
\usepackage[accsupp]{axessibility}  

\definecolor{cvprblue}{rgb}{0.21,0.49,0.74}
\usepackage[pagebackref,breaklinks,colorlinks,allcolors=cvprblue]{hyperref}

\def\paperID{13} 
\def\confName{CVPR}
\def\confYear{2025}

\title{Visualizing and Controlling Cortical Responses Using Voxel-Weighted Activation Maximization}

\author{Matthew W. Shinkle\\
Department of Cognitive and Brain Sciences\\
University of Nevada, Reno\\
{\tt\small matthewshinkle@gmail.com}
\and
Mark D. Lescroart\\
Department of Cognitive and Brain Sciences\\
University of Nevada, Reno\\
{\tt\small mlescroart@unr.edu}
}

\begin{document}
\maketitle

\begin{abstract}
Deep neural networks (DNNs) trained on visual tasks develop feature representations that resemble those in the human visual system. Although DNN-based encoding models can accurately predict brain responses to visual stimuli, they offer limited insight into the specific features driving these responses. Here, we demonstrate that activation maximization—a technique designed to interpret vision DNNs—can be applied to DNN-based encoding models of the human brain. We extract and adaptively downsample activations from multiple layers of a pretrained Inception V3 network, then use linear regression to predict fMRI responses. This yields a full image-computable model of brain responses. Next, we apply activation maximization to generate images optimized for predicted responses in individual cortical voxels. We find that these images contain visual characteristics that qualitatively correspond with known selectivity and enable exploration of selectivity across the visual cortex. We further extend our method to whole regions of interest (ROIs) of the brain and validate its efficacy by presenting these images to human participants in an fMRI study. We find that the generated images reliably drive activity in targeted regions across both low- and high-level visual areas and across subjects. These results demonstrate that activation maximization can be successfully applied to DNN-based encoding models. By addressing key limitations of alternative approaches that require natively generative models, our approach enables flexible characterization and modulation of responses across the human visual system.
\end{abstract}
    
\section{Introduction}
\label{sec:intro}

Deep neural networks (DNNs) trained on vision tasks generate internal representations that closely resemble features extracted by the human visual system~\cite{lecunDeepLearning2015, kriegeskorteDeepNeuralNetworks2015}. DNNs learn hierarchical representations—from simple edge detectors to complex object representations—that parallel stages of processing in visual cortex~\cite{yaminsPerformanceoptimizedHierarchicalModels2014}. However, an overall similarity between representations in DNNs and brains reveals little about detailed selectivity in either system.

One way to leverage DNNs to extract interpretable insights about the brain is by generating images designed to drive responses in specific brain regions. Previous methods have used evolutionary algorithms~\cite{ponceEvolvingImagesVisual2019}, deconvolutional networks~\cite{gucluDeepNeuralNetworks2015}, generative adversarial networks (GANs)~\cite{guNeuroGenActivationOptimized2022}, and diffusion models~\cite{luoBrainDiffusionVisual2023} to visualize selectivity in the brain. However, these methods have limitations: closed-loop genetic algorithms require extensive repeated measurements, and GAN- or diffusion-based approaches rely on natively generative models, which restricts their application to a small set of model architectures and training objectives.

Activation maximization offers a promising alternative, as it generates images using pretrained, non-generative DNN models~\cite{Olah2017}. Originally designed to characterize internal features within DNNs, activation maximization iteratively modifies an input image to maximize the activation of a targeted unit or set of units, thereby revealing the features to which the network is most sensitive~\cite{Olah2017}. This raises the question of whether activation maximization can be used to produce interpretable visualizations of selectivity in the brain. Previous work by Bashivan \etal~\cite{bashivanNeuralPopulationControl2019} applied activation maximization to encoding models using activations of a single DNN layer and generated images that drove electrophysiological responses in macaque V4. However, the extent to which this approach generalizes to fMRI recordings in humans remains unexplored. Moreover, restricting models to individual DNN layers may limit their ability to capture complex selectivity across the human visual system.

In this work, we demonstrate that activation maximization can be applied to DNN-based encoding models to characterize feature selectivity across the human visual system. Specifically, we apply adaptive downsampling to activations drawn from layers across a pretrained DNN, train an encoding model to predict fMRI responses based on downsampled DNN activations, and then use activation maximization to generate images optimized for predicted cortical responses. We then present a subset of these images to human participants and analyze their fMRI responses.

We report three primary findings. First, activation maximization applied via our approach generates interpretable synthetic images for single voxels and whole ROIs. Second, these images effectively capture detailed visual selectivity that aligns with known functional selectivity across visual cortex. Finally, images optimized for predicted ROI responses reliably drive fMRI responses in targeted regions.

To our knowledge, this work is the first demonstration that activation maximization can be successfully applied to whole-network DNN-based encoding models to characterize and modulate fMRI responses across the human visual system. Furthermore, our approach enables exploration of selectivity within and beyond established cortical regions.
\section{Methods}
\label{sec:methods}

Our approach consists of two main stages: (1) constructing a DNN-based encoding model that accurately predicts fMRI responses to visual stimuli, and (2) applying activation maximization to generate synthetic images that drive responses in different cortical regions.

\begin{figure}[htbp]
    \centering
    \includegraphics[width=\columnwidth]{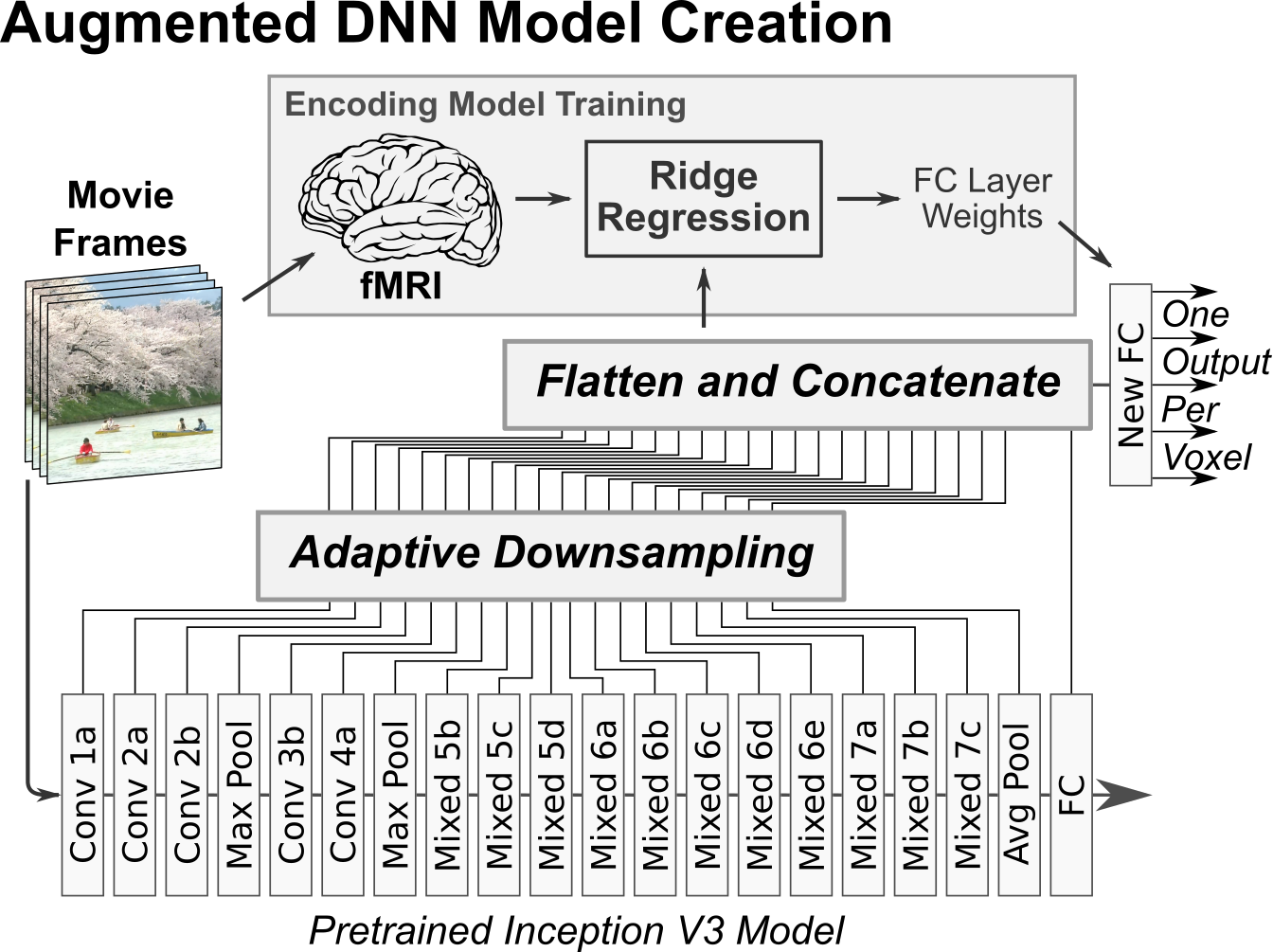}
    \caption{\textbf{Model Overview.} Schematic of our augmented DNN framework. Movie frames are processed by a pretrained Inception V3 model. Activations are extracted from multiple layers and adaptively downsampled, flattened, and concatenated to form a feature vector that is passed through a new fully connected layer. The weights of this layer are trained via ridge regression to predict voxelwise fMRI responses.}
    \label{fig:model}
\end{figure}

\textbf{Core Encoding Model}  
Training a randomly initialized deep neural network to predict fMRI responses from image inputs is infeasible given limits on the amount of fMRI data that can be reasonably collected relative to the number of trainable parameters. Instead, we leverage the fact that DNNs trained on visual tasks develop internal representations that closely resemble those in human visual cortex~\cite{lecunDeepLearning2015, kriegeskorteDeepNeuralNetworks2015}. We use Inception V3~\cite{szegedyRethinkingInceptionArchitecture2015}, pretrained on ImageNet for object classification, as our backbone model. Using a large set of naturalistic movie clips with corresponding fMRI responses~\cite{huthContinuousSemanticSpace2012}, we extract activations in response to each video frame from Inception V3's convolutional layers, inception blocks, and final fully connected layer. This yields millions of features per stimulus, so we apply an adaptive spatial downsampling strategy to reduce each feature map to a fixed size—approximately 5,000 features per layer—roughly balancing feature counts across layers while preserving spatial information. The downsampled activations are then flattened and concatenated into approximately 78,000 features per frame. We then use ridge regression to fit a voxelwise encoding model~\cite{naselarisEncodingDecodingFMRI2011} based on these features. The result is a predictive model mapping visual inputs to brain responses. As shown in Supplementary Results, this model accurately predicts fMRI responses across the visual system, including in all areas examined in our experiments.

\textbf{Activation Maximization Algorithm}  
To translate our encoding model into interpretable visualizations of cortical selectivity, we apply activation maximization (or feature visualization)~\cite{erhan2009visualizing, Nguyen2016, Olah2017}. We conceptualize the ridge regression weights as forming a new fully connected (FC) layer appended to the Inception V3 model. Starting from an initial image of random noise, the augmented model predicts voxel responses; backpropagation then computes gradients with respect to the input image, revealing modifications that increase the predicted response. This iterative process—repeated for 2,500 iterations and incorporating techniques such as random image augmentations and optimization in Fourier space~\cite{Olah2017}—yields images optimized for predicted responses in individual voxels or regions.

\textbf{Experimental Design}  
We evaluate our approach through three experiments. In \textbf{Experiment 1}, we generate synthetic images that maximize predicted responses of individual voxels, and compare these images to known visual selectivity. In \textbf{Experiment 2}, we generate images for entire cortical regions by averaging predicted responses across voxels, and then present these images to human participants undergoing fMRI to assess whether they drive the targeted responses. Finally, in \textbf{Experiment 3}, we evaluate the generalizability of our approach by measuring responses to images synthesized for a separate set of subjects.
\section{Results}
\label{sec:results}

\subsection{Experiment 1: Individual Voxel Visualizations}

\begin{figure}[htbp]
    \centering
    \includegraphics[width=\columnwidth]{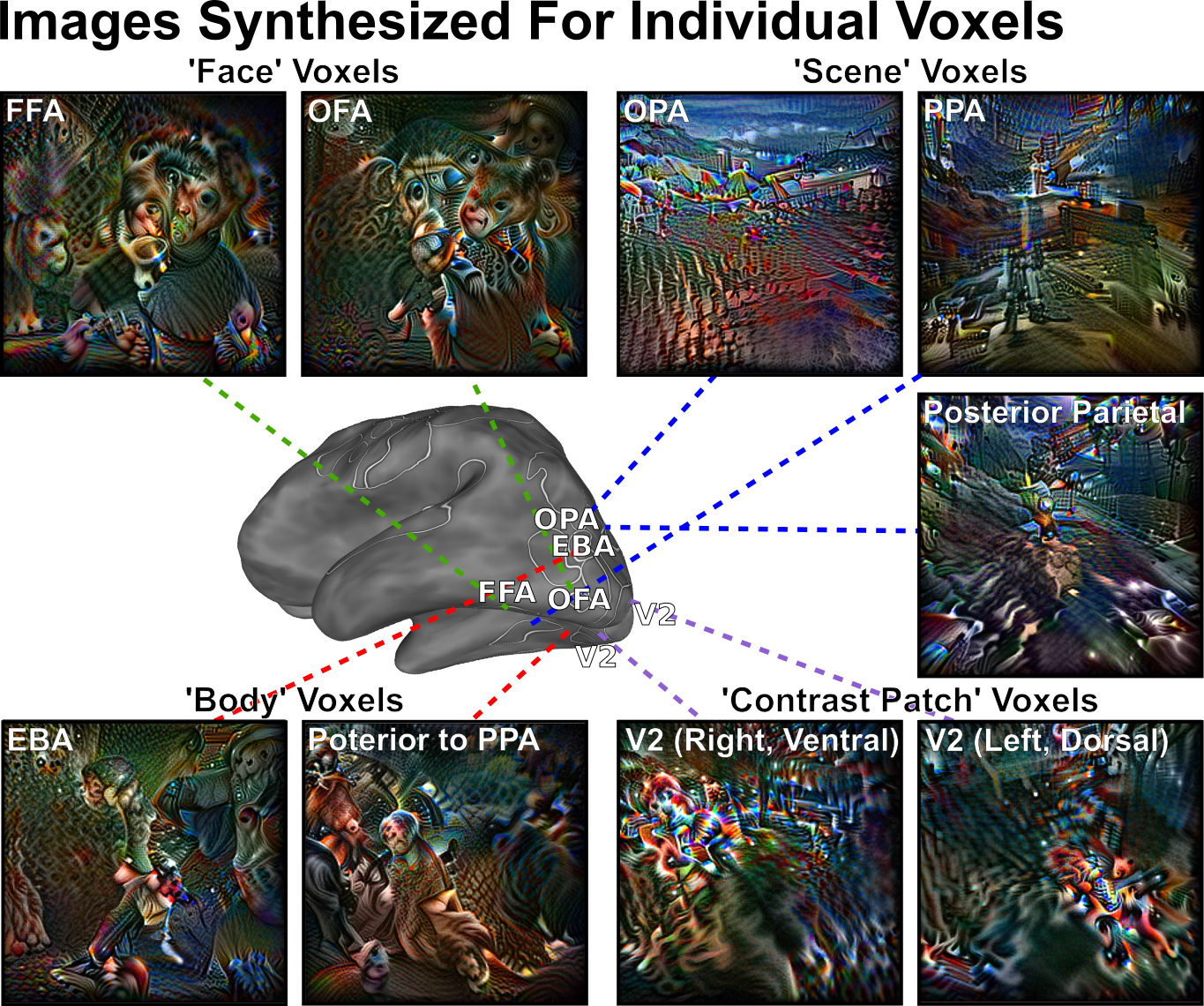}
    \caption{\textbf{Synthetic images optimized for individual cortical voxels.} Images synthesized for individual voxels in V2 show high-contrast patches in expected retinotopic locations; those for voxels drawn from face-selective regions (OFA/FFA) exhibit facial features, including eye- and face-like components; and those from scene-selective regions (OPA) display scene-like elements. In addition, images synthesized for voxels outside of established cortical regions confirm the presence of an additional scene-selective patch in the posterior parietal lobe and show preliminary evidence of body-selective voxels directly posterior to PPA. Images are best viewed zoomed in.}
    \label{fig:individual_voxels}
\end{figure}

We used voxel-weighted activation maximization to synthesize images based on predicted responses of individual cortical voxels. These images can be explored interactively at \url{http://piecesofmind.psyc.unr.edu/pycortex/activation_maximization/}. Shown in ~\cref{fig:individual_voxels}, the contents of these images qualitatively match the known selectivity of the regions from which the voxels were drawn. Images generated for voxels in dorsal V2 in the left hemisphere contained high-contrast patches localized in the lower right visual field, whereas voxels from ventral V2 in the right hemisphere yielded patches in the upper left visual field. These results correspond well with the known retinotopic selectivity in early visual cortex; indeed, these images strongly resemble stimuli designed to elicit robust responses in retinotopic mapping studies~\cite{Hansen2007}.

In contrast, images synthesized for individual voxels in face-selective regions (OFA, FFA) exhibit clear facial features, particularly eyes and mouths. Images from scene-selective voxels contain elements resembling buildings, streets, and ground planes receding in depth. Finally, we note two findings based on images synthesized for voxels outside of conventional ROIs. First, we find a patch of voxels in posterior parietal lobe with apparent scene-selectivity, corresponding to a purported scene-selective area recently discovered in this location~\cite{b.kennedyPreviouslyUndescribedSceneselective2024}. Second, we note a patch of voxels posterior to PPA, for which synthesized images contain characteristics akin to upright bodies, similar to images generated for a known body-selective area (EBA). These findings demonstrate the exploratory value of our approach and suggest directions for future research.

\subsection{Experiment 2: Empirical Evaluation of ROI-Optimized Images}

\begin{figure*}[htbp]
    \centering
    \includegraphics[width=\textwidth]{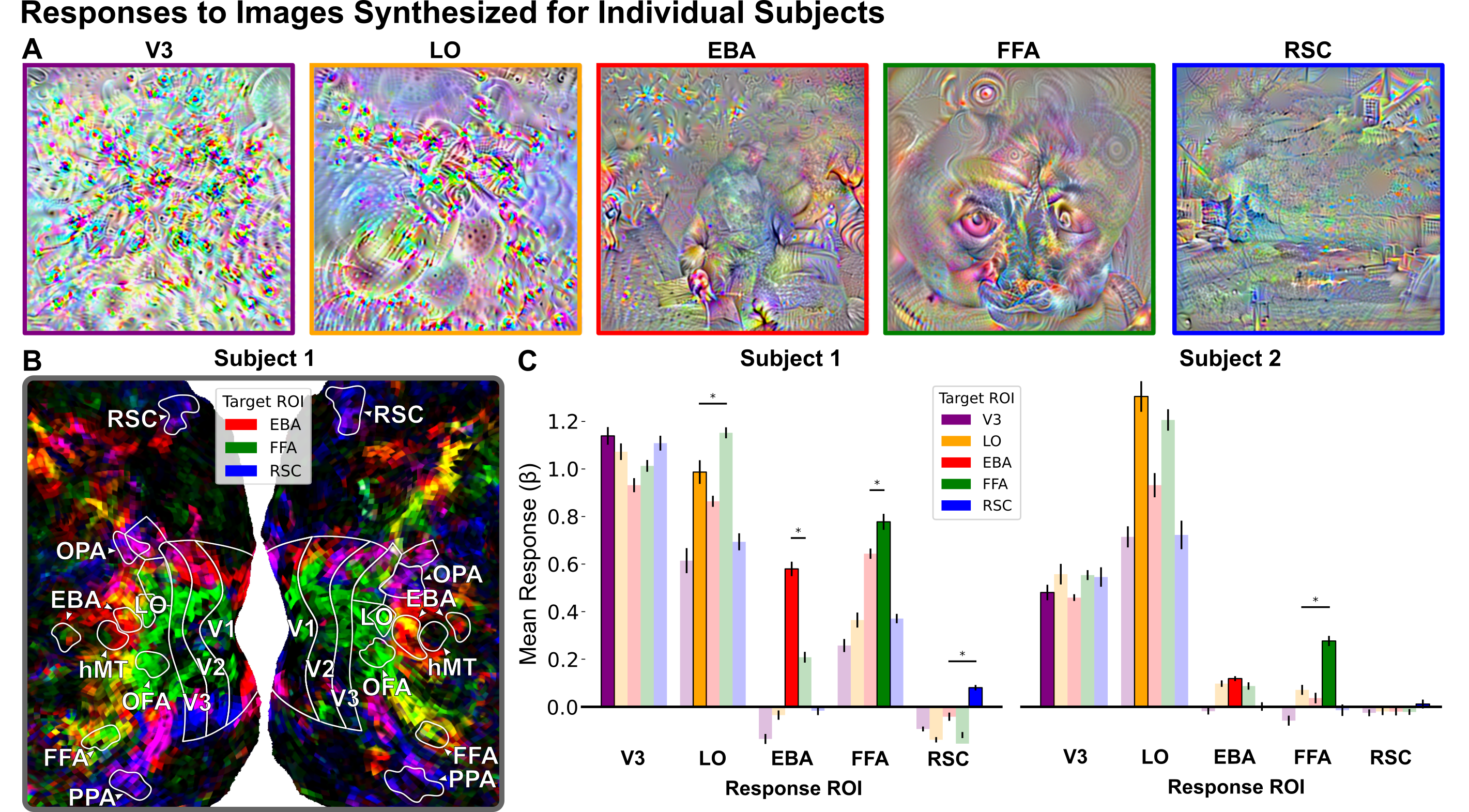}
    \caption{\textbf{Real fMRI responses to images synthesized for whole ROIs.} \textbf{(A)} Synthetic images optimized via activation maximization for five target regions (V3, LO, FFA, EBA, and RSC) from Subject 1. These images contain characteristics which qualitatively correspond to known region selectivity. For example, V3 images exhibit high-contrast patterns spanning the visual field (consistent with the relatively low-level visual feature selectivity in V3), images for FFA contain recognizable elements of faces, and images optimized for RSC appear scene-like. \textbf{(B)} fMRI response contrasts to images optimized for EBA (red), FFA (green), and RSC (blue) shown on the cortical surface of subject 1. Voxels within these regions are consistently most strongly driven by images optimized for them. \textbf{(C)} ROI-wise bar plots display fMRI response magnitudes to images optimized for different ROI targets. Fully opaque bars show the response evoked by images expressly optimized for that same ROI. For both subjects, the highlighted response is higher than the others in four out of five ROIs, indicating that the synthetic images not only qualitatively match known selectivity but also effectively drive cortical responses.}
    \label{fig:exp2}
\end{figure*}

To evaluate whether images synthesized via our approach actually drove responses in these regions, we first aggregated weights across voxels in each region, and then optimized images based on predicted whole-region responses. ~\cref{fig:exp2} illustrates the synthetic images optimized for five target ROIs (V3, LO, FFA, EBA, and RSC). These images reveal region-specific features: V3 images contain patterns of high contrast across the visual field, face-selective ROIs display recognizable facial elements, and images for scene-selective areas exhibit naturalistic scene-like contents. Next, we presented these images to human subjects and collected real fMRI responses. Shown in ~\cref{fig:exp2}B, contrast response maps reveal differential activation patterns across the cortical surface when images are optimized for EBA, FFA, or RSC. The ROI-wise bar plots (~\cref{fig:exp2}C) further indicate that, in four out of five ROIs, the response evoked by images optimized specifically for that ROI is higher than responses evoked by images optimized for other regions, with this difference being significant in EBA, FFA, and RSC for subject 1 and FFA for subject 2 (independent-samples t-test, df = 18, t(18) $>$ 2.101, p $<$ .05). This validates that the synthetic images effectively drove region-specific fMRI responses.

\subsection{Experiment 3: Cross-Subject Generalization}
To assess whether synthesized images drive responses across subjects, we generated synthetic images for the same five ROIs as before based on the average predicted response across a set of four subjects. We then presented these images to a separate set of four subjects. We found that these images robustly drove fMRI responses in targeted ROIs across subjects, with similar response magnitudes to those used in Experiment 2. Across subjects, ROI responses were significantly higher for images optimized for that same ROI in V3, EBA, and FFA (independent-samples t-test, df = 18, t(18) $>$ 2.101, p $<$ .05), and non-significantly so in RSC. This demonstrates that our approach, and the images it generates, capture aspects of region-specific selectivity that generalize well across individuals. See Supplementary Results for visualizations and details of these results.
\section{Conclusions and Discussion}

In this work, we introduced and evaluated a technique that leverages the representations of deep neural networks to synthesize images that characterize and control fMRI responses in the brain. In Experiment 1, we found that synthesizing images predicted to drive individual voxels reveals region-specific retinotopic and categorical selectivity. In Experiment 2, our ROI-level synthesis demonstrated that the generated images effectively drive targeted fMRI responses. Finally, in Experiment 3, we showed that these synthesized images evoke consistent responses across subjects. This indicates that our method captures general tuning properties of regions rather than subject-specific tuning. Combined, these results suggest that the synthesized images provide meaningful reflections of cortical tuning and lend credence to exploratory findings in other regions.

In one such exploration, we observed a patch of apparent scene-selective voxels in the posterior parietal lobe that aligns with recent findings from Kennedy \etal~\cite{b.kennedyPreviouslyUndescribedSceneselective2024}. Additionally, our observation of body-like selectivity in voxels posterior to the PPA is novel and warrants experimental validation. We did not exhaustively explore all voxels outside established regions in this work; examination of images for other voxels may reveal additional, yet-unidentified functional areas. Moreover, optimizing images for individual voxels and subregions within ROIs could further illuminate their fine-grained functional organization.

Importantly, our method addresses key limitations of previous approaches. By eliminating the need for natively generative models~\cite{guNeuroGenActivationOptimized2022,luoBrainDiffusionVisual2023}, our method could be used to visualize cortical selectivity via DNN models with diverse architectures and training objectives. Furthermore, activation maximization works with different optimization objectives, which enables optimization for—or against—any differentiable image property. Finally, by drawing activations from all layers of a model, our approach leverages the full range of internal representations a model has learned.

While we empirically evaluated the effects of synthesized images on fMRI responses, our characterization of their contents remains qualitative. Quantification of image features in the synthesized images would enable a more objective assessment of selectivity of the regions or voxels for which they were optimized. Finally, generation of other stimulus formats—such as videos or text—and application to other measurement techniques, such as magnetoencephalography (MEG), would extend our approach and enable fuller characterization of cortical selectivity.
\section*{Acknowledgements}
Funding for this work has been provided by NIH COBRE awd-01-00002341 and a 2023–2024 Nevada Space Grant Consortium Fellowship awarded to Matthew Shinkle.

{
    \small
    \bibliographystyle{ieeenat_fullname}
    \bibliography{main}

\begin{thebibliography}{21}
\providecommand{\natexlab}[1]{#1}
\providecommand{\url}[1]{\texttt{#1}}
\expandafter\ifx\csname urlstyle\endcsname\relax
  \providecommand{\doi}[1]{doi: #1}\else
  \providecommand{\doi}{doi: \begingroup \urlstyle{rm}\Url}\fi

\bibitem[Bashivan et~al.(2019)Bashivan, Kar, and DiCarlo]{bashivanNeuralPopulationControl2019}
Pouya Bashivan, Kohitij Kar, and James~J. DiCarlo.
\newblock Neural population control via deep image synthesis.
\newblock \emph{Science}, 364\penalty0 (6439):\penalty0 eaav9436, 2019.

\bibitem[Dupré La~Tour et~al.(2022)Dupré La~Tour, Eickenberg, Nunez-Elizalde, and Gallant]{duprelatourFeaturespaceSelectionBanded2022}
Tom Dupré La~Tour, Michael Eickenberg, Anwar~O. Nunez-Elizalde, and Jack~L. Gallant.
\newblock Feature-space selection with banded ridge regression.
\newblock \emph{NeuroImage}, 264:\penalty0 119728, 2022.

\bibitem[Erhan et~al.(2009)Erhan, Bengio, Courville, and Vincent]{erhan2009visualizing}
Dumitru Erhan, Yoshua Bengio, Aaron Courville, and Pascal Vincent.
\newblock Visualizing higher-layer features of a deep network.
\newblock \emph{University of Montreal}, 1341\penalty0 (3):\penalty0 1, 2009.

\bibitem[Gao et~al.(2015)Gao, Huth, Lescroart, and Gallant]{Gao2015}
James~S. Gao, Alexander~G. Huth, Mark~D. Lescroart, and Jack~L. Gallant.
\newblock Pycortex: an interactive surface visualizer for fmri.
\newblock \emph{Frontiers in Neuroinformatics}, 9, 2015.

\bibitem[Gu et~al.(2022)Gu, Jamison, Khosla, Allen, Wu, St-Yves, Naselaris, Kay, Sabuncu, and Kuceyeski]{guNeuroGenActivationOptimized2022}
Zijin Gu, Keith~Wakefield Jamison, Meenakshi Khosla, Emily~J. Allen, Yihan Wu, Ghislain St-Yves, Thomas Naselaris, Kendrick Kay, Mert~R. Sabuncu, and Amy Kuceyeski.
\newblock {NeuroGen}: Activation optimized image synthesis for discovery neuroscience.
\newblock \emph{NeuroImage}, 247:\penalty0 118812, 2022.

\bibitem[Guclu and Van~Gerven(2015)]{gucluDeepNeuralNetworks2015}
U. Guclu and M.~A.~J. Van~Gerven.
\newblock Deep neural networks reveal a gradient in the complexity of neural representations across the ventral stream.
\newblock \emph{Journal of Neuroscience}, 35\penalty0 (27):\penalty0 10005--10014, 2015.

\bibitem[Hansen et~al.(2007)Hansen, Kay, and Gallant]{Hansen2007}
Kathleen~A. Hansen, Kendrick~N. Kay, and Jack~L. Gallant.
\newblock Topographic organization in and near human visual area v4.
\newblock \emph{The Journal of Neuroscience}, 27\penalty0 (44):\penalty0 11896--11911, 2007.

\bibitem[He et~al.(2014)He, Zhang, Ren, and Sun]{He_2014}
Kaiming He, Xiangyu Zhang, Shaoqing Ren, and Jian Sun.
\newblock \emph{Spatial Pyramid Pooling in Deep Convolutional Networks for Visual Recognition}, pages 346--361.
\newblock Springer International Publishing, 2014.

\bibitem[Huth et~al.(2012)Huth, Nishimoto, Vu, and Gallant]{huthContinuousSemanticSpace2012}
Alexander~G. Huth, Shinji Nishimoto, An~T. Vu, and Jack~L. Gallant.
\newblock A continuous semantic space describes the representation of thousands of object and action categories across the human brain.
\newblock \emph{Neuron}, 76\penalty0 (6):\penalty0 1210--1224, 2012.

\bibitem[Kennedy et~al.(2024)Kennedy, Malladi, Tootell, and Nasr]{b.kennedyPreviouslyUndescribedSceneselective2024}
B. Kennedy, Sarala~N. Malladi, R. Tootell, and S. Nasr.
\newblock A previously undescribed scene-selective site is the key to encoding ego-motion in naturalistic environments.
\newblock \emph{eLife}, 2024.

\bibitem[Kriegeskorte(2015)]{kriegeskorteDeepNeuralNetworks2015}
Nikolaus Kriegeskorte.
\newblock Deep neural networks: A new framework for modeling biological vision and brain information processing.
\newblock \emph{Annual Review of Vision Science}, 1\penalty0 (1):\penalty0 417--446, 2015.

\bibitem[LeCun et~al.(2015)LeCun, Bengio, and Hinton]{lecunDeepLearning2015}
Yann LeCun, Yoshua Bengio, and Geoffrey Hinton.
\newblock Deep learning.
\newblock \emph{Nature}, 521\penalty0 (7553):\penalty0 436--444, 2015.

\bibitem[Luo et~al.(2023)Luo, Henderson, Wehbe, and Tarr]{luoBrainDiffusionVisual2023}
Andrew~F. Luo, Margaret~M. Henderson, Leila Wehbe, and Michael~J. Tarr.
\newblock Brain diffusion for visual exploration: Cortical discovery using large scale generative models, 2023.

\bibitem[Naselaris et~al.(2011)Naselaris, Kay, Nishimoto, and Gallant]{naselarisEncodingDecodingFMRI2011}
Thomas Naselaris, Kendrick~N. Kay, Shinji Nishimoto, and Jack~L. Gallant.
\newblock Encoding and decoding in fmri.
\newblock \emph{NeuroImage}, 56\penalty0 (2):\penalty0 400--410, 2011.

\bibitem[Nguyen et~al.(2016)Nguyen, Yosinski, and Clune]{Nguyen2016}
Anh Nguyen, Jason Yosinski, and Jeff Clune.
\newblock Multifaceted feature visualization: Uncovering the different types of features learned by each neuron in deep neural networks, 2016.

\bibitem[Olah et~al.(2017)Olah, Mordvintsev, and Schubert]{Olah2017}
Chris Olah, Alexander Mordvintsev, and Ludwig Schubert.
\newblock Feature visualization.
\newblock \emph{Distill}, 2\penalty0 (11), 2017.

\bibitem[Ponce et~al.(2019)Ponce, Xiao, Schade, Hartmann, Kreiman, and Livingstone]{ponceEvolvingImagesVisual2019}
Carlos~R. Ponce, Will Xiao, Peter~F. Schade, Till~S. Hartmann, Gabriel Kreiman, and Margaret~S. Livingstone.
\newblock Evolving images for visual neurons using a deep generative network reveals coding principles and neuronal preferences.
\newblock \emph{Cell}, 177\penalty0 (4):\penalty0 999--1009.e10, 2019.

\bibitem[Popham et~al.(2021)Popham, Huth, Bilenko, Deniz, Gao, Nunez-Elizalde, and Gallant]{Popham2021}
Sara~F Popham, Alexander~G Huth, Natalia~Y Bilenko, Fatma Deniz, James~S Gao, Anwar~O Nunez-Elizalde, and Jack~L Gallant.
\newblock Visual and linguistic semantic representations are aligned at the border of human visual cortex.
\newblock \emph{Nature Neuroscience}, 24\penalty0 (11):\penalty0 1628--1636, 2021.

\bibitem[Prince et~al.(2022)Prince, Charest, Kurzawski, Pyles, Tarr, and Kay]{Prince2022}
Jacob~S Prince, Ian Charest, Jan~W Kurzawski, John~A Pyles, Michael~J Tarr, and Kendrick~N Kay.
\newblock Improving the accuracy of single-trial fmri response estimates using glmsingle.
\newblock \emph{eLife}, 11, 2022.

\bibitem[Szegedy et~al.(2015)Szegedy, Vanhoucke, Ioffe, Shlens, and Wojna]{szegedyRethinkingInceptionArchitecture2015}
Christian Szegedy, Vincent Vanhoucke, Sergey Ioffe, Jonathon Shlens, and Zbigniew Wojna.
\newblock Rethinking the inception architecture for computer vision, 2015.

\bibitem[Yamins et~al.(2014)Yamins, Hong, Cadieu, Solomon, Seibert, and DiCarlo]{yaminsPerformanceoptimizedHierarchicalModels2014}
Daniel L.~K. Yamins, Ha Hong, Charles~F. Cadieu, Ethan~A. Solomon, Darren Seibert, and James~J. DiCarlo.
\newblock Performance-optimized hierarchical models predict neural responses in higher visual cortex.
\newblock \emph{Proceedings of the National Academy of Sciences}, 111\penalty0 (23):\penalty0 8619--8624, 2014.

\end{thebibliography}
}

\clearpage
\setcounter{page}{1}
\maketitlesupplementary

\section{Supplementary Results}
\label{sec:supp_results}

\begin{figure*}[htbp]
    \centering
    \includegraphics[width=\textwidth]{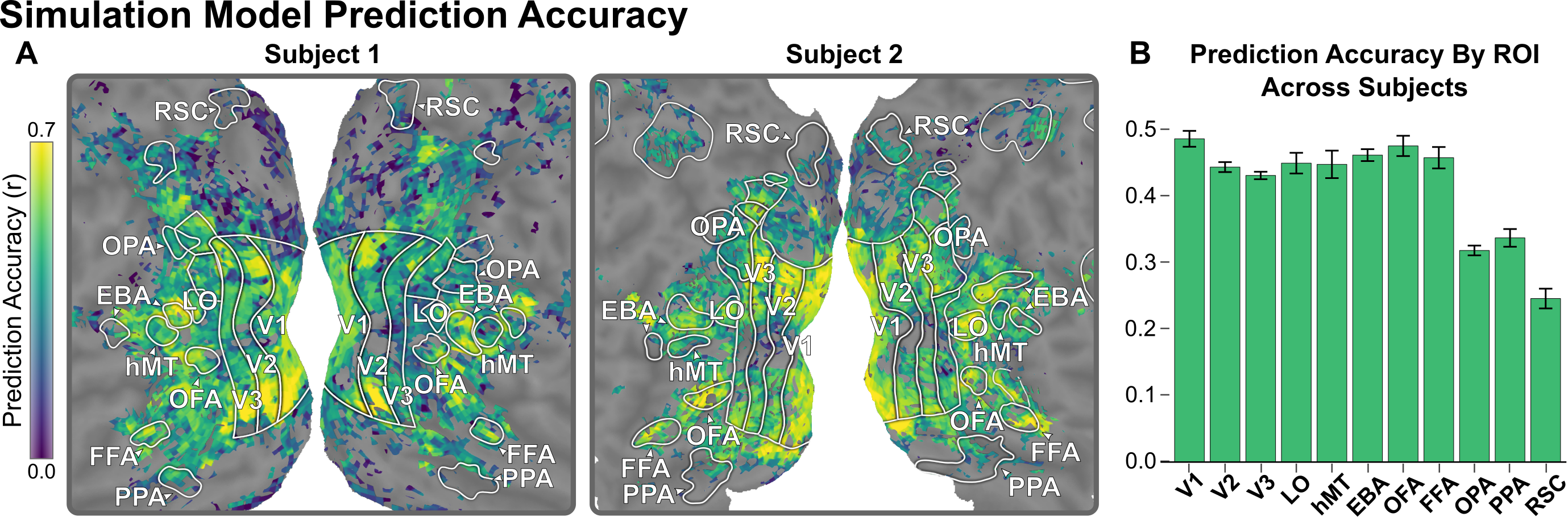}
    \caption{\textbf{Prediction Accuracy of the Inception V3-Based Encoding Model.} \textbf{(A)} Flatmaps of voxelwise prediction accuracy for subjects 1 and 2 in natural movie data, illustrating the encoding model prediction accuracy across the visual cortex. \textbf{(B)} Bar chart displaying ROI-wise prediction accuracy, noise-ceiling corrected and averaged across voxels and subjects, for early and higher-level regions. Error bars show SEM across subjects.}
    \label{fig:prediction_accuracy}
\end{figure*}

To evaluate the overall quality of the DNN-to-brain mapping, we computed the prediction accuracy of our Inception V3-based encoding model on the testing set of naturalistic movie clips. Prediction accuracy was noise-ceiling corrected and averaged across voxels and subjects for each region of interest (ROI). Shown in \cref{fig:prediction_accuracy}A and B, our results reveal high prediction accuracy across a wide range of visual cortical areas. High prediction accuracy in early visual areas (V1, V2, V3) suggests that our model captures selectivity to low-level visual features such as contrast and edges. High prediction accuracy in higher-level regions, including face-selective areas (OFA and FFA) and the body-selective Extrastriate Body Area (EBA), demonstrates that the same model also captures selectivity for more high-level visual features. Additionally, although slightly lower, prediction accuracy is consistently high in scene-selective regions such as the Occipital Place Area (OPA), Parahippocampal Place Area (PPA), and the Retrosplenial Complex (RSC). These results demonstrate that—by drawing from layers across Inception V3—a single encoding model can effectively model responses across visual areas. 

Furthermore, \cref{fig:prediction_accuracy}A shows that this model also predicts responses accurately in many voxels outside of established regions. Taken together, these results show that our model can effectively predict fMRI responses to diverse visual characteristics across the visual system. This serves as a critical foundation for predicting fMRI responses to new stimuli, including our synthetic stimuli.

\subsection{Experiment 3: Closed-Loop Optimization and Cross-Subject Generalization}

\begin{figure*}[htbp]
    \centering
    \includegraphics[width=\textwidth]{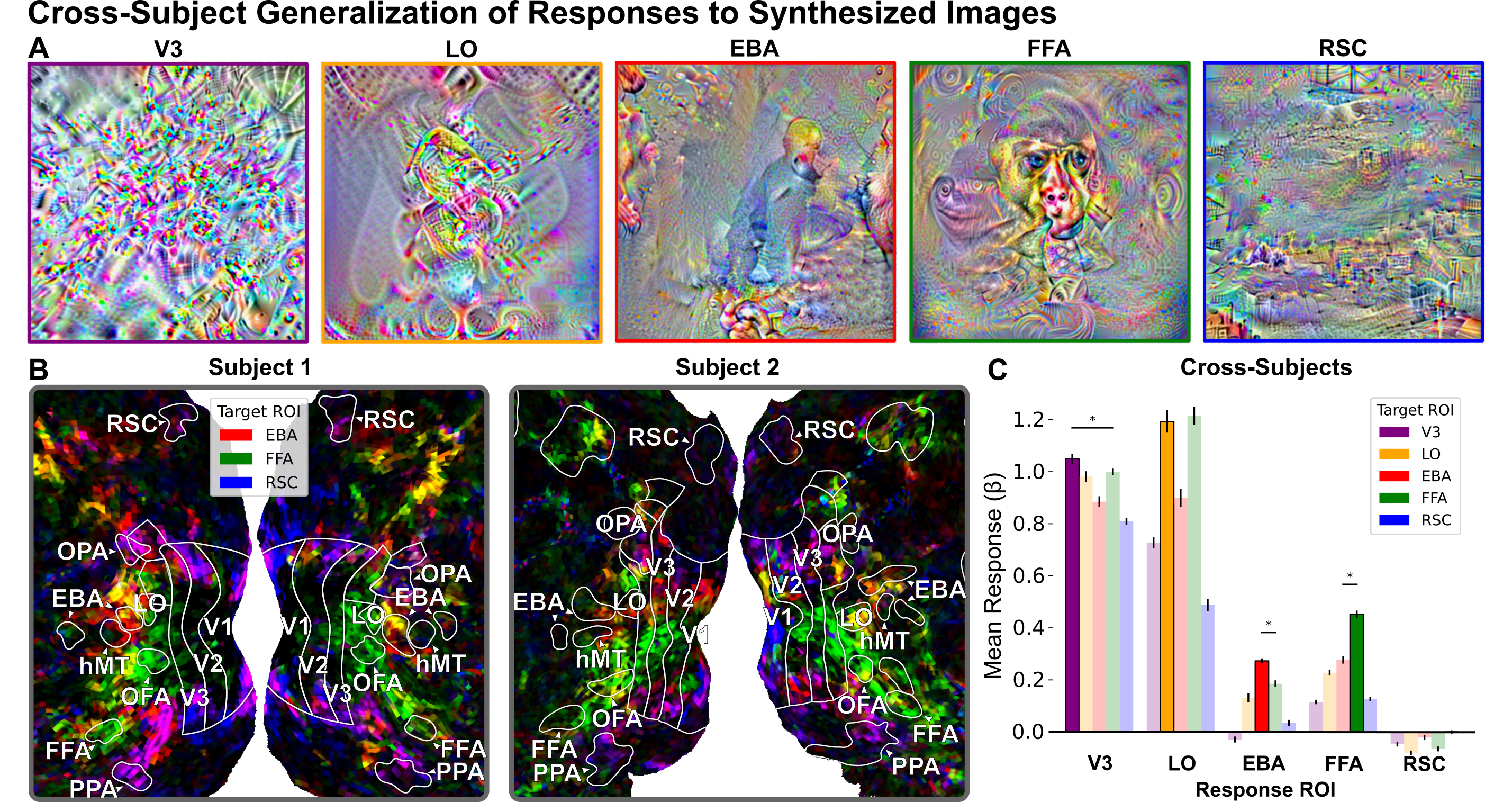}
    \caption{\textbf{Cross-Subject Generalization of ROI-Level Synthetic Images.} \textbf{(A)} Example synthetic images for each ROI, generated using predicted responses from a group of four subjects and then presented to a separate set of four subjects. These images capture region-specific features consistent with known selectivity. \textbf{(B)} Flatmap visualizations of contrast responses for subjects 1 and 2 illustrate the spatial distribution of activation differences evoked by the synthetic images. \textbf{(C)} A single bar plot shows the average fMRI response across subjects for the synthetic images, with the highlighted bars corresponding to responses evoked by images optimized for each target ROI. Together, these results demonstrate robust cross-subject generalization of our voxel-weighted activation maximization framework.}
    \label{fig:cross_subject_ROIs}
\end{figure*}

~\cref{fig:cross_subject_ROIs} summarizes the key findings from the cross-subject analysis. ~\cref{fig:cross_subject_ROIs}A displays example synthetic images for each target ROI (V3, LO, FFA, EBA, and RSC) generated from the averaged weights. These images exhibit region‐specific features consistent with known selectivity—for example, clear facial elements in face-selective regions and natural scene-like patterns in scene-selective areas. ~\cref{fig:cross_subject_ROIs}B presents flatmap visualizations of the contrasted responses for images optimized for EBA (red), FFA (green), and RSC (blue) for Subjects 1 and 2, demonstrating that the spatial distribution of the evoked responses aligns with the expected anatomical boundaries. Finally, ~\cref{fig:cross_subject_ROIs}C shows ROI-wise bar plots of the average fMRI responses across the four test subjects. In each ROI, the response to images optimized for that ROI was higher than the responses to images optimized for other regions in four out of five ROIs, thereby confirming that the synthetic images robustly drive the intended cortical responses across subjects.

These results indicate that our voxel-weighted activation maximization framework is not subject-specific but captures features that generalize across individuals. The robust activation patterns observed in the cross-subject evaluation further support the utility of our method in probing the functional selectivity of cortical regions.

\section{Supplementary Methods}
\label{sec:supp_methods}

\subsection{fMRI Data Acquisition and Preprocessing}
For fitting our DNN-based encoding models, we used BOLD fMRI responses to a large set of naturalistic movie clips from Huth \etal~\cite{huthContinuousSemanticSpace2012}. This consisted of video clips depicting a wide variety of dynamic scenes, including people, animals, objects, and natural environments. These stimuli were divided into 120 minutes of training clips and 9 minutes of testing clips. We used BOLD responses to these stimuli for 6 subjects (4 male, 2 female); data for five of these subjects was previously collected and data for one subject was collected for this study. During scanning, participants maintained fixation on a central fixation cross. See Huth \etal~\cite{huthContinuousSemanticSpace2012} and Popham \etal~\cite{Popham2021} for more details.

Voxels were filtered based on a noise ceiling criterion that estimates response reliability across repeated stimulus presentations. Specifically, the noise ceiling was computed as the mean pairwise correlation between BOLD responses to repeated presentations of the naturalistic testing clips. Voxels with a noise ceiling significantly above zero (p $<$ 0.05, uncorrected) were included in subsequent analyses.

In addition, for experiments 2 and 3 we collected BOLD responses to images synthesized via our activation maximization approach. We collected BOLD data for four subjects (2 male, 2 female) using a Siemens Skyra 3T scanner with a 32-channel head coil. Blood oxygenation level dependent (BOLD) fMRI data were collected with a repetition time (TR) of 1000 ms, an echo time (TE) of 31 ms, and a voxel resolution of 3.2 × 3.2 × 2.6 mm. Each scanning run consisted of 100 stimulus images, preceded by 12 blank trials, followed by 16 blank trials, and interleaved with 10 additional blank (50\% luminance gray screen) trials to ensure robust baseline estimation and to minimize adaptation effects. During BOLD data collection, participants were tasked with fixating on a central fixation dot while attending to the image contents.

Estimation of responses to each image was performed using GLMsingle~\cite{Prince2022}. Standard preprocessing procedures (\eg, motion correction) were applied to all BOLD data. Functional data for experiments 2 and 3 were temporally downsampled to an effective TR of 2.0 seconds to reduce physiological noise. 

Experimental protocols were approved by the Institutional Review Boards of the University of Nevada, Reno and the University of California, Berkeley.

\subsection{DNN Activation Extraction and Downsampling}
\textbf{Image Preprocessing:}  
Naturalistic movie frames and synthesized images were preprocessed using the standard transformations expected by the Inception V3 model. Each image was resized to 299 pixels, center-cropped to produce a 299×299 image, and normalized using the ImageNet mean values [0.485, 0.456, 0.406] and standard deviations [0.229, 0.224, 0.225]. These operations are fully differentiable, which allows us to optimize higher-resolution images (\eg, 500×500 pixels) despite the network receiving inputs at a resolution of 299×299.

\textbf{Layer Selection:}  
Activations were extracted from 21 layers of the Inception V3 network~\cite{szegedyRethinkingInceptionArchitecture2015}:
\begin{enumerate}
    \item Conv 1a (3×3)
    \item Conv 2a (3×3)
    \item Conv 2b (3×3)
    \item MaxPool 1
    \item Conv 3b (1×1)
    \item Conv 4a (3×3)
    \item MaxPool 2
    \item Inception 5b
    \item Inception 5c
    \item Inception 5d
    \item Reduction 6a
    \item Inception 6b
    \item Inception 6c
    \item Inception 6d
    \item Inception 6e
    \item Reduction 7a
    \item Inception 7b
    \item Inception 7c
    \item AvgPool
    \item Dropout
    \item FC
\end{enumerate}

This selection spans from low-level features to high-level semantic representations.

\textbf{Adaptive Spatial Downsampling:}  
To manage the high dimensionality of activations from convolutional layers, we employ adaptive spatial pooling. We use PyTorch’s adaptive pooling functions (similar to those described in He \etal~\cite{He_2014}) to reduce each feature map to a fixed output size such that approximately 5,000 features are retained per layer. The target output size for each spatial dimension is computed by

\begin{equation}
  S_i = \left\lfloor \left(\frac{F_{\text{max}}}{C}\right)^{\frac{1}{n}} \right\rfloor
  \label{eq:Si}
\end{equation}

where \(F_{\text{max}} = 5000\), \(C\) is the number of channels, and \(n\) is the number of spatial dimensions. Since \(n\) can vary, this formulation generalizes beyond 2D and is applicable to higher-dimensional feature maps (\eg, those from 3D CNNs).

\textbf{Temporal Downsampling:}  
Due to the mismatch between stimulus framerate (15 HZ) and fMRI repetition time (0.5 HZ), we average the downsampled features extracted from frames of the naturalistic movie clips over each 2-second window to match the temporal resolution of the BOLD signal.

\subsection{Voxelwise Encoding Model Fitting}
In order to generate a predictive mapping from extracted DNN activations and fMRI responses, we fit voxelwise encoding models~\cite{naselarisEncodingDecodingFMRI2011}. To do this, we first extract downsampled layerwise activations (see above) to the training set of naturalistic movie clips. These downsampled activations from each layer are flattened and concatenated, yielding a feature vector of roughly 78,000 elements per stimulus. We then fit a ridge regression model to predict voxelwise fMRI responses from these features. Prior to regression, the fMRI responses are z-scored across TRs, and temporal lags of 2, 4, and 6 seconds are combined into a single design matrix.

For each voxel \(i\), the regression model is formulated as

\begin{equation}
  \hat{\beta}_i = \left( \mathbf{X}^\top \mathbf{X} + \alpha I \right)^{-1} \mathbf{X}^\top \mathbf{y}_i
  \label{eq:beta_hat}
\end{equation}

where \(\mathbf{X}\) is the design matrix containing the concatenated DNN features, \(\mathbf{y}_i\) is the z-scored fMRI response vector for voxel \(i\), \(\alpha\) is the regularization parameter, and \(I\) is the identity matrix. The regularization parameter (\(\alpha\)) is selected from a set of 15 values that are logarithmically interpolated between \(10^{0}\) and \(10^{10}\). For each voxel, the optimal \(\alpha\) is chosen based on maximizing the cross-validated \(R^2\) using 10 splits and 10 resamplings (n\_splits = 10, n\_resamps = 10). Regression fitting was performed using the tikreg package~\cite{duprelatourFeaturespaceSelectionBanded2022}.

\subsection{Activation Maximization}

We perform activation maximization by optimizing an input image to maximize the predicted fMRI response in a target voxel or region. The optimization is carried out in the Fourier domain, which biases the solution towards smooth, interpretable patterns and avoids high-frequency artifacts~\cite{Olah2017}. 

\begin{figure}[htbp]
    \centering
    \includegraphics[width=\columnwidth]{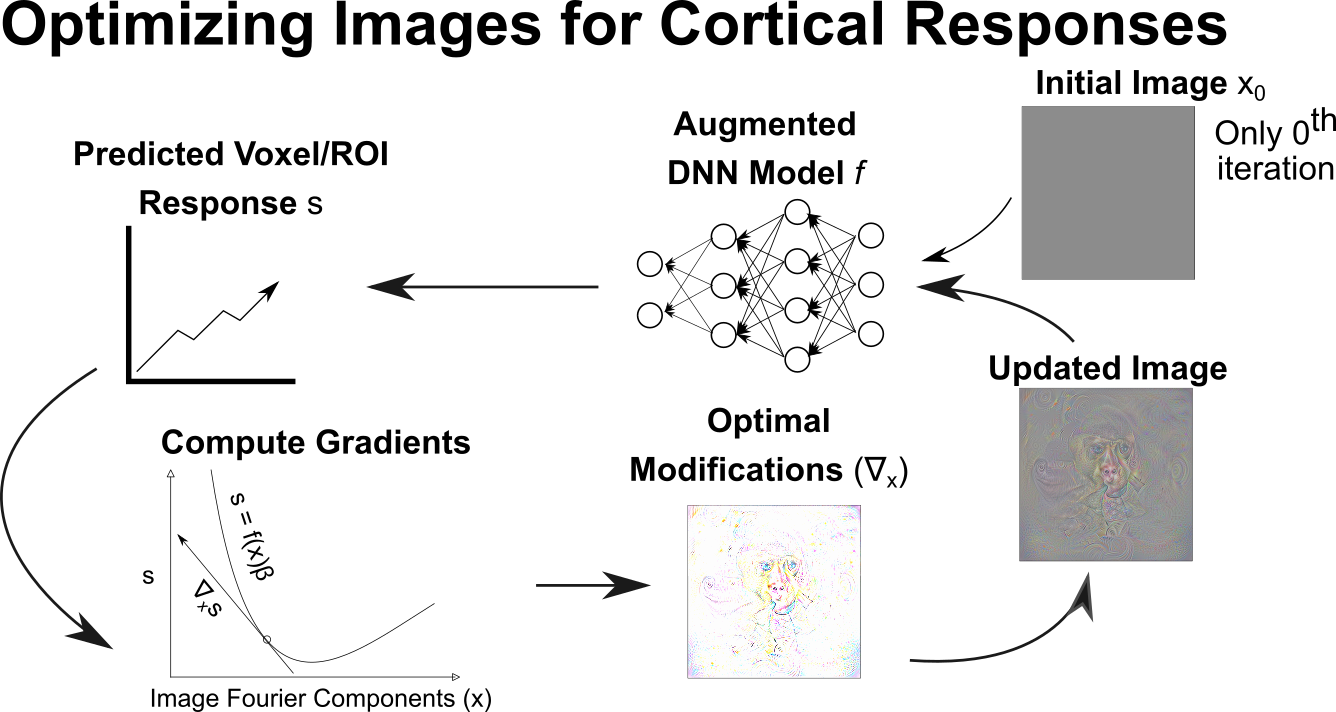}
    \caption{\textbf{Generation Process.} Diagram illustrating the activation maximization procedure. Starting from an initial neutral grayscale image in the Fourier domain, the image is transformed into the spatial domain using an inverse FFT. Random invariance transforms (\eg, cropping, rotation within \(-5^\circ\) to \(5^\circ\), and resized cropping) are applied to regularize the optimization. The preprocessed image is then fed through the augmented network, and gradients are computed and backpropagated (using the Adam optimizer with a learning rate of \(1 \times 10^{-2}\)) to update the Fourier coefficients. This iterative gradient ascent process is repeated for 2,500 iterations to synthesize an image that maximizes the predicted fMRI response.}
    \label{fig:generation}
\end{figure}

The procedure is as follows:
\begin{itemize}
    \item \textbf{Initialization:} The input is initialized as a neutral grayscale image (with a pixel value of 140) in Fourier space.
    \item \textbf{Inverse Fourier Transform:} The image is transformed to the spatial domain via an inverse FFT, taking the magnitude to obtain real-valued pixel intensities.
    \item \textbf{Image Augmentations:} To regularize optimization, we apply a series of random invariance transformations:
    \begin{itemize}
        \item A random crop to 500×500 pixels with a padding of 5 pixels.
        \item A random rotation between \(-5^\circ\) and \(5^\circ\).
        \item A random resized crop to 500×500 pixels with a scaling factor between 0.95 and 1.05, maintaining a 1:1 aspect ratio.
        \item A second random crop to 500×500 pixels with a padding of 3 pixels.
    \end{itemize}
    \item \textbf{Preprocessing:} The augmented image is resized to 299×299 pixels, center-cropped, and normalized using the standard Inception V3 transformations.
    \item \textbf{Forward Pass and Loss Computation:} The preprocessed image is passed through the augmented network to obtain downsampled activations. The predicted fMRI response is computed as the dot product between the flattened activations and the target weight vector. We define the loss as the negative predicted response:
    \begin{equation}
      \mathcal{L}(\mathbf{x}) = -\left( \mathbf{f}(\mathbf{x}) \cdot \boldsymbol{\beta} \right)
      \label{eq:loss1}
    \end{equation}
    \item \textbf{Gradient Update:} Gradients are backpropagated to the Fourier domain representation, and the Adam optimizer (with a learning rate of \(1 \times 10^{-2}\)) updates the Fourier coefficients. The optimization runs for 2,500 iterations.
\end{itemize}

For synthetic images optimized for single-voxel responses, we incorporated two modifications to enhance interpretability. First, the image initialization was set to be completely black with a small amount of noise added. We found that this initialization yields a clearer distinction between the emergent image content and the background. Second, we applied a color channel decorrelation step based on the approach described in Olah \etal~\cite{Olah2017}. This decorrelation reduces redundancy across the red, green, and blue channels, encouraging the emergence of distinct color features in the synthesized images. While these modifications improve the qualitative interpretability of the generated images, their effects on actual fMRI responses remain a subject for future investigation.

\subsection{Optimization Objectives}
For all three experiments, we create an optimization objective by computing contrasts in the space of regression weights. This contrastive approach minimizes the impact of shared visual selectivity among voxels or regions while amplifying their differences. Concretely, let \(\boldsymbol{\beta} \in \mathbb{R}^d\) denote the regression weight vector for a given voxel or region. First, we normalize the weights by performing z-scoring across features:

\begin{equation}
  \mathbf{z} = \frac{\boldsymbol{\beta} - \mu(\boldsymbol{\beta})}{\sigma(\boldsymbol{\beta})}
  \label{eq:z}
\end{equation}

where \(\mu(\boldsymbol{\beta})\) and \(\sigma(\boldsymbol{\beta})\) are the mean and standard deviation of \(\boldsymbol{\beta}\), respectively. This normalizes the scale of the weights (which is influenced by factors such as fMRI signal strength).

Next, we subtract an estimate of the mean feature selectivity—computed either across all cortical voxels (for individual voxel optimization) or across the five selected regions (for ROI-level optimization)—from the z-scored weights:

\begin{equation}
  \boldsymbol{\beta}_{\text{contrast}} = \mathbf{z} - \bar{\mathbf{z}}
  \label{eq:beta_contrast}
\end{equation}

with \(\bar{\mathbf{z}}\) representing the average z-scored weight vector. This step effectively removes common components of visual selectivity across voxels, allowing the optimization to focus on features that are uniquely pronounced in the target voxel or region.

Finally, to ensure a consistent rate of image optimization during activation maximization, we normalize the contrast weights by their L2 norm:

\begin{equation}
  \boldsymbol{\beta}_{\text{final}} = \frac{\boldsymbol{\beta}_{\text{contrast}}}{\|\boldsymbol{\beta}_{\text{contrast}}\|}
  \label{eq:beta_final}
\end{equation}

This stabilizes the gradient ascent process by ensuring that the optimization target has a consistent scale, independent of the magnitude of the regression weights.

During image optimization, the normalized contrast weights are multiplied by the feature vector \(\mathbf{f}(\mathbf{x})\) derived from the DNN layers, yielding an estimated contrast value:

\begin{equation}
  s = \boldsymbol{\beta}_{\text{final}}^\top \mathbf{f}(\mathbf{x})
  \label{eq:s}
\end{equation}

This estimated contrast value is then used as the optimization target. Specifically, we define the loss function as

\begin{equation}
  \mathcal{L}(\mathbf{x}) = -s
  \label{eq:loss2}
\end{equation}

and apply gradient descent to minimize \(\mathcal{L}(\mathbf{x})\), which is equivalent to performing gradient ascent on \(s\).

\subsection{Software and Implementation}
Regression fitting was performed using the tikreg package~\cite{duprelatourFeaturespaceSelectionBanded2022}. Beta estimation of the fMRI response was conducted using GLMsingle~\cite{Prince2022} and ROIs and cortical surface visualizations were generated using Pycortex~\cite{Gao2015}. Custom PyTorch code for DNN feature extraction and synthetic image generation is available at \url{https://github.com/MShinkle/VWAM}.

\end{document}